\documentclass[conference]{IEEEtran}
\usepackage{graphicx}
\usepackage{amsmath}
\usepackage{amssymb}

\usepackage{amsmath,amsfonts}
\usepackage{array}
\usepackage{algorithm,algorithmic}
\usepackage{textcomp}
\usepackage{url}
\usepackage{graphicx}
\usepackage{cite}
\usepackage[hidelinks]{hyperref}
\usepackage{eso-pic}

\begin{document}

\AddToShipoutPicture*{
\AtPageUpperLeft{
\raisebox{-0.8cm}{   
\hspace{0.8cm}       
\parbox{0.9\textwidth}{
\small
2025 28th International Conference on Computer and Information Technology (ICCIT)\\
19--21 December 2025, Cox’s Bazar, Bangladesh
}
}
}
}

\AddToShipoutPicture*{
\AtPageLowerLeft{
\raisebox{1cm}{
\hspace{0.8cm}
\parbox{0.9\textwidth}{
\small
979-8-3315-7867-1/25/\$31.00~\copyright~2025 IEEE
}
}
}
}

\title{Resource-Aware Evolutionary Neural Architecture Search for Cardiac MRI Segmentation}

\author{Farhana Yasmin\,\href{https://orcid.org/0000-0002-5247-0958}{0000-0002-5247-0958}$^{1,3}$, Mahade Hasan\,\href{https://orcid.org/0009-0006-2778-1216}{0009-0006-2778-1216}$^{2,3*}$,\\ Haipeng Liu\,\href{https://orcid.org/0000-0002-4212-2503}{0000-0002-4212-2503}$^{4,5}$, Amjad Ali\,\href{https://orcid.org/0000-0002-5346-1017}{0000-0002-5346-1017}$^{6}$, \\ Ghulam Muhammad\,\href{https://orcid.org/0000-0002-9781-3969}{0000-0002-9781-3969}$^{7}$, Yu Xue\,\href{https://orcid.org/0000-0002-9069-7547}{0000-0002-9069-7547}$^{2}$\\ \\ \textsuperscript{1}\textit{School of Computer Science, Nanjing University of Information Science and Technology, Nanjing, China.} \\ \textsuperscript{2}\textit{School of Software, Nanjing University of Information Science and Technology, Nanjing, China.} \\ \textsuperscript{3}\textit{Eastern University, Ashulia Model Town, Dhaka, 1345, Bangladesh.}\\ \textsuperscript{4}\textit{Research Centre for Intelligent Healthcare, Coventry University, Coventry CV1 5RW, UK.} \\ \textsuperscript{4}\textit{National Medical Research Association, Leicester, UK.} \\ \textsuperscript{6}\textit{Faculty of Engineering and Technology, Muscat University, Oman.} \\ \textsuperscript{7}\textit{College of Computer and Information Sciences, King Saud University, Riyadh, Saudi Arabia.} \\ 
\textit{Corresponding Author: Mahade Hasan (mhasan@nuist.edu.cn)}

}


\maketitle

\begin{abstract}
Cardiac magnetic resonance (CMR) segmentation underpins quantitative assessment of ventricular structure and function, yet reliable delineation remains difficult due to low tissue contrast, fuzzy boundaries, and inter-scan variability. We present CardiacNAS, an evolutionary neural architecture search (NAS) framework that couples a U-Net–like supernet with a cardiac-aware search space spanning depth/width, kernel size, filter size, attention, fusion, activation, dropout, and residual scaling. The search is explicitly resource-aware, jointly optimizing dice similarity coefficient (DSC) and 95th-percentile Hausdorff distance (HD95) versus model size and floating-point operations (FLOPs) under fixed compute budgets. Candidate architectures are instantiated from the supernet, trained with proxy budgets, and evolved through crossover, mutation, and elitist selection. We evaluate on the ACDC dataset and compare against six state-of-the-art methods, using qualitative comparisons, learning-curve analyses, and design-factor correlation studies. The resulting model attains 93.22\% average DSC and 4.73,mm HD95 with 3.58M parameters and 14.56 GFLOPs, demonstrating a favorable accuracy-efficiency trade-off. Analyses indicate that searched attention and fusion choices, together with residual scaling, contribute to improved boundary fidelity and stability. CardiacNAS offers a principled, resource-aware approach to deployable CMR segmentation with transparent reporting of architectural complexity and compute budgets.
\end{abstract}


\section{Introduction}
Cardiac magnetic resonance (CMR) segmentation enables quantitative assessment of ventricular anatomy and function by delineating the left ventricle (LV), right ventricle (RV), and myocardium (MYO) \cite{aghapanah2025mecardnet}\cite{qin2024urca}. Reliable contours are challenging due to low tissue contrast, fuzzy boundaries around papillary muscles and insertion points, inter-patient variability, and domain shift across scanners and protocols \cite{10530285}\cite{10821880}. These factors make boundary accuracy and efficiency central to downstream clinical use, where dice similarity coefficient (DSC) and the 95th-percentile Hausdorff distance (HD95) are standard indicators of region overlap and contour fidelity \cite{10423891}\cite{EvoGrayNet}.

Deep learning has become the dominant paradigm for medical image segmentation, with fully convolutional and encoder–decoder architectures forming the backbone \cite{ren2025hresformer}\cite{hasan2023rule}\cite{10587153}\cite{yasmin2025developing}. Successive designs add multi-scale context, denser skip connections, attention, and residual pathways \cite{huang2025learnable}\cite{zhu2025merging}\cite{10771659} \cite{hasan2025advances}. Despite progress, practitioners often face trade-offs between boundary quality and computational efficiency, which complicates deployment on resource-limited systems. Neural architecture search (NAS) aims to automate design under such constraints, spanning differentiable, evolutionary, and hybrid strategies \cite{yasmin2026epso}\cite{YU2025101837}\cite{ZECHEN2025128155}\cite{HU2025128338}. However, many pipelines still require substantial engineering to balance accuracy, FLOPs, and parameter counts. 

\begin{figure*}[h]
\centerline{\includegraphics[width=19cm,height=10cm]{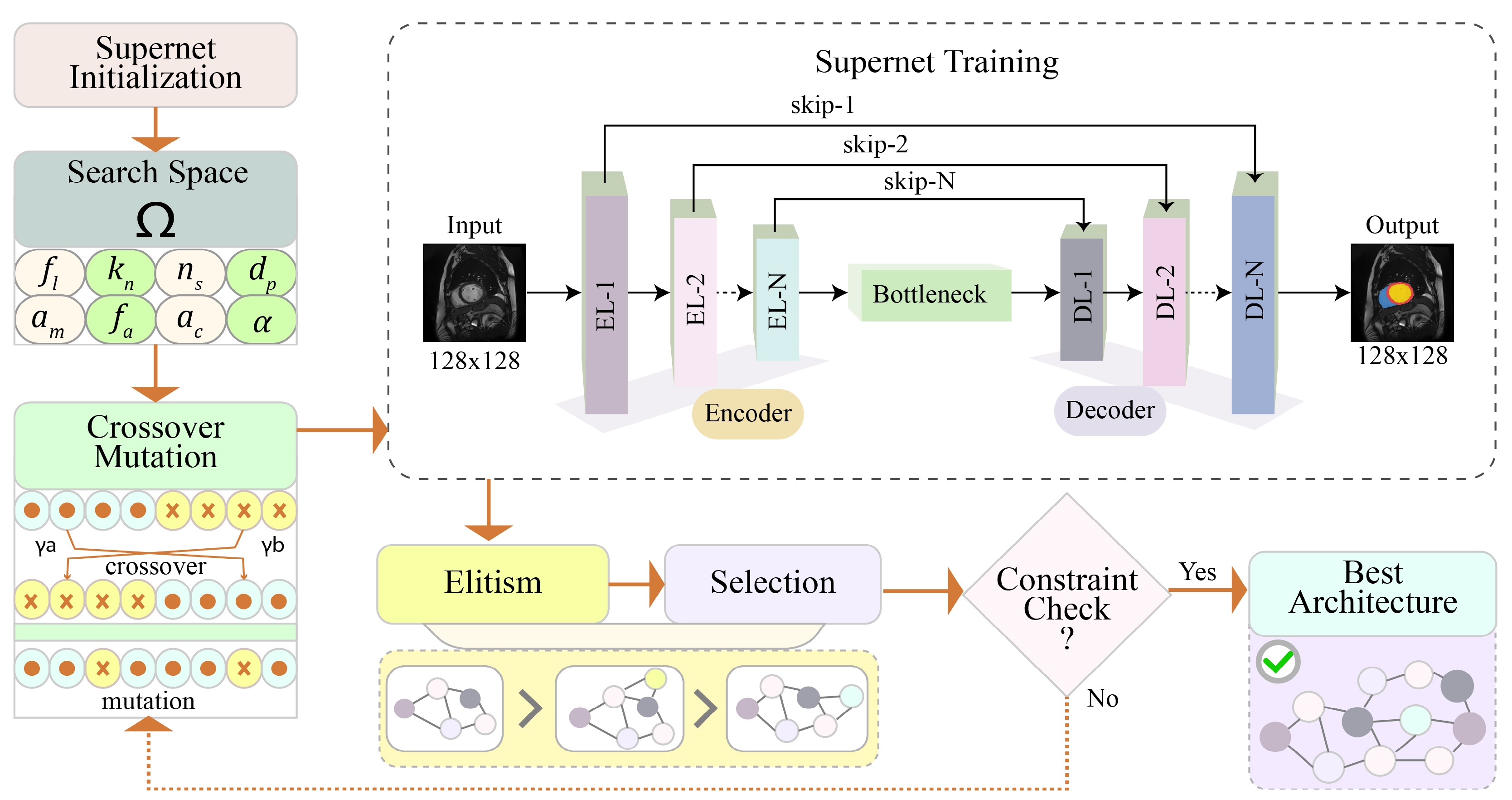}}
\caption{Overall framework of CardiacNAS, showing supernet initialization, candidate generation through crossover and mutation, and evolutionary selection steps leading to the final architecture.}

 \label{Fig_1}
\end{figure*} 

To address these limitations, we present \emph{CardiacNAS}, an evolutionary NAS framework designed for CMR segmentation while considering both accuracy and efficiency.

The key contributions are as follows:
\begin{itemize}
    \item We propose a cardiac-aware supernet and search space that integrate multi-scale context, attention, and flexible skip-fusion to target crisp LV/MYO/RV boundaries under tight complexity budgets.
    \item We propose an evolutionary NAS procedure with proxy training and resource-aware selection that explicitly trades off segmentation quality (DSC/HD95) against model complexity (parameters/FLOPs).
    \item We provide a comprehensive evaluation, including qualitative comparisons, learning-curve analyses, and design-factor correlation studies that clarify how attention, fusion, and residual scaling influence boundary fidelity and stability.
\end{itemize}

The rest of the paper is structured as follows: Section~\ref{sec:related works} reviews medical image segmentation and NAS methods; Section~\ref{method} details the CardiacNAS framework; Section~\ref{result} reports the experimental setup, results, and analyses; and Section~\ref{conclusion} summarizes the findings and outlines future work.

\section{Related Works}
\label{sec:related works}

This section summarizes prior work on medical image segmentation and NAS-based segmentation frameworks.

\subsection{Medical Image Segmentation}
Medical image segmentation enables quantitative analysis by delineating organs and lesions across MRI, CT, ultrasound, and endoscopy. Encoder–decoder CNNs remain central, beginning with U\!-Net \cite{unet2015} and extending to designs that enrich multi-scale context and efficiency including FTransDeepLab \cite{feng2025ftransdeeplab} and the self-configuring nnU\!-Net \cite{isensee2021nnu}. Hybrid CNN–Transformer models such as TransUNet \cite{TransUNet} capture long-range dependencies while preserving local inductive biases. In contrast to hand-crafted backbones and single-objective setups, \textit{CardiacNAS} couples a U\!-Net–like supernet with a cardiac-aware search space (depth/width, kernel size and dilation, attention, fusion, activation, dropout, residual scaling) and uses evolutionary search under explicit accuracy–efficiency constraints (DSC/HD95 vs.\ parameters/FLOPs), aligning with trends toward multi-objective, resource-aware design.

\subsection{NAS-Based Segmentation Approaches}
NAS reduces manual design effort and systematically explores accuracy–efficiency trade-offs. Representative strategies include multi-objective cell/backbone search, such as MOE-FewSeg \cite{11025554} and evolutionary search over compact operator/topology spaces, such as BiX\!-NAS \cite{10009}. Recent medical NAS frameworks expand to hybrid CNN–Transformer operators with efficient (single-/multi-path) supernets, including HCT\!-Net \cite{Yu2023} and M-ENAS \cite{Li2024}. In contrast, CardiacNAS adopts a resource-aware NAS formulation balancing segmentation fidelity (DSC/HD95) against model complexity (Params/FLOPs) under explicit compute budgets and documented learning dynamics, enabling deployable cardiac segmentation.

\section{Methods}
\label{method}
The proposed CardiacNAS framework integrates a supernet architecture with an evolutionary neural architecture search to design efficient models for cardiac MRI segmentation. As shown in Fig.~\ref{Fig_1}, the workflow begins with defining a supernet and a parameterized search space that controls depth, filter size, kernel size, attention modules, and fusion strategies. Candidate architectures are first generated randomly and then progressively improved through evolutionary operations such as crossover and mutation. Each candidate is trained with a limited budget and evaluated on segmentation accuracy and efficiency measures, while constraint checks ensure feasibility. Through repeated selection and elitism across generations, CardiacNAS converges toward high-performing architectures, from which the final model is chosen.

\subsection{Supernet Architecture}
We employ a U-Net like supernet as the backbone of CardiacNAS, structured into an encoder, bottleneck, and decoder with skip connections. This framework provides the search space within which candidate architectures are instantiated and evolved.

\subsubsection{Input} Each cardiac MR slice is denoted as $x \in \mathbb{R}^{H \times W \times 1}$, normalized prior to training. The network processes fixed-size inputs ($128 \times 128$).

\begin{table}[t]
\caption{\textsc{\small Search Space Parameters Used for Architecture Generation in CardiacNAS}}
\small
\setlength{\tabcolsep}{5pt}
\begin{tabular}{p{100pt}p{125pt}}
\hline

\textbf{Variable} & \textbf{Search Space} \\
\hline
Filter sizes ($f_l$) & [32 - 127] \\
Kernel size ($k_n$) & [1 - 7] \\
Number of layers ($n_s$) & [2 - 4] \\
Dropout rate ($d_p$) & [0.1 - 0.5] \\
Attention modules ($a_m$) & [squeeze-excitation, self-attention] \\
Fusion/aggregation ($f_a$) & [add, concat, weighted sum] \\
Activation functions ($a_c$) & [relu, elu, tanh, sigmoid] \\
Residual scaling factor ($\alpha$) & continuous [0.1 – 1.0] \\
\hline
\end{tabular}
\label{tab:random_arch}
\end{table}

\begin{algorithm}[h]
\fontsize{8}{9}\selectfont
\caption{CardiacNAS: Neural Architecture Search}
\label{alg:cardiacnas_eas}
\begin{algorithmic}[1]

\STATE \textbf{Input:} training set $\mathcal{D}_{tr}$, validation set $\mathcal{D}_{va}$, search space $\Omega$, population size $M$, generations $T$, crossover rate $\pi_\chi$, mutation rate $\pi_\mu$
\STATE \textbf{Output:} best architecture $\hat{\gamma}$

\STATE \textit{// Initialization}
\STATE Sample initial population $\mathcal{G}^{(0)}=\{\gamma_1,\ldots,\gamma_M\}$ with $\gamma_i \sim \Omega$ \quad \textit{(genotypes over variables in Table~\ref{tab:random_arch})}
\STATE $\hat{\gamma} \leftarrow \varnothing$, \; $\hat{f} \leftarrow -\infty$

\FOR{$t=1$ \TO $T$}
    \STATE \textit{// Parent selection (e.g., tournament)}
    \STATE Build mating pool $\mathcal{M}$ by repeatedly selecting parents from $\mathcal{G}^{(t-1)}$

    \STATE \textit{// Variation: crossover \& mutation}
    \STATE $\mathcal{O} \leftarrow \emptyset$ \quad \textit{(offspring)}
    \WHILE{$|\mathcal{O}| < M$}
        \STATE Pick $(\gamma^a,\gamma^b)$ from $\mathcal{M}$
        \IF{$\mathrm{rand}()<\pi_\chi$}
            \STATE $\tilde{\gamma} \leftarrow \mathsf{UniformCrossover}(\gamma^a,\gamma^b)$ \textit{// per-gene swap; blend for continuous $\alpha$}
        \ELSE
            \STATE $\tilde{\gamma} \leftarrow \gamma^a$
        \ENDIF
        \IF{$\mathrm{rand}()<\pi_\mu$}
            \STATE $\tilde{\gamma} \leftarrow \mathsf{Mutate}(\tilde{\gamma};\Omega)$ \textit{// resample one/few genes: $f_l,k_n,n_s,d_p,a_m,f_a,a_c,\alpha$}
        \ENDIF
        \STATE $\mathcal{O} \leftarrow \mathcal{O} \cup \{\tilde{\gamma}\}$
    \ENDWHILE

    \STATE \textit{// Evaluation (proxy training + validation)}
    \FOR{each $\gamma \in \mathcal{O}$}
        \STATE Build model $\mathcal{M}_\gamma$ from genotype $\gamma$; train on $\mathcal{D}_{tr}$ with early stopping/warm start
        \STATE Compute fitness $f(\gamma)$ on $\mathcal{D}_{va}$ (e.g., Dice $\uparrow$ with complexity penalty)
        \IF{$f(\gamma)>\hat{f}$} \STATE $\hat{\gamma}\leftarrow \gamma$, \; $\hat{f}\leftarrow f(\gamma)$ \ENDIF
    \ENDFOR

    \STATE \textit{// Environmental selection (elitist replacement)}
    \STATE $\mathcal{G}^{(t)} \leftarrow \mathsf{ElitistReplace}(\mathcal{G}^{(t-1)}, \mathcal{O}, M)$
\ENDFOR

\STATE \textbf{return} $\hat{\gamma}$

\end{algorithmic}
\end{algorithm}
\subsubsection{Encoder} The encoder extracts hierarchical features through convolutional blocks parameterized by the genotype $\gamma = (f_l, k_n, n_s, d_p, a_m, f_a, a_c, \alpha)$. For stage $i$, the feature maps are given by
\[
\mathbf{F}_i = \phi_i \big( \mathcal{C}(\mathbf{F}_{i-1}; f_l^i, k_n^i) \big), \quad i=1,\ldots,n_s, \tag{1}
\]
Here, $\mathcal{C}$ denotes the convolution operation, $\phi_i$ is the activation $a_c$, and dropout $d_p$ is applied optionally. Attention modules $a_m$ may be inserted to refine channel or spatial responses.

\subsubsection{Bottleneck} At the lowest resolution, multi-scale features are aggregated:
\[
\mathbf{F}_b = \sum_{r \in \mathcal{R}} \mathcal{C}(\mathbf{F}_{n_s}; f_l^b, k_n^b, d_r=r), \tag{2}
\]
Here $\mathcal{R}$ is the set of dilation rates. This stage can include advanced recalibration or attention depending on $\gamma$.

\subsubsection{Decoder} The decoder reconstructs spatial detail using upsampling and fusion. For stage $j$, the feature map is
\[
\mathbf{U}_j = \psi_j \big( \mathcal{U}(\mathbf{U}_{j-1}) \; \oplus_{f_a} \; \mathbf{F}_{n_s-j+1} \big), \tag{3}
\]
Here $\mathcal{U}$ is an upsampling operation, $\oplus_{f_a}$ denotes fusion through strategy $f_a \in \{\text{add}, \text{concat}, \text{weighted sum}\}$, and $\psi_j$ is the decoder block function. Residual scaling is applied as
\[
\mathbf{U}_j \leftarrow \alpha \cdot \mathbf{U}_j + (1-\alpha)\cdot \mathbf{F}_{n_s-j+1}, \tag{4}
\]

\subsubsection{Output} The final segmentation map is obtained as
\[
\hat{y} = \text{Softmax}\big(\mathcal{C}(\mathbf{U}_{n_s}; C, 1)\big), \tag{5}
\]
Here $C$ is the number of classes (background, LV, MYO, RV). This supernet formulation ensures that CardiacNAS can flexibly explore architectures through $\gamma$, while preserving a consistent encoder–decoder backbone designed for cardiac MRI segmentation.

\subsection{Evolutionary Search}
We design the evolutionary search process in CardiacNAS around three core stages: defining the search space, applying crossover and mutation operations, and conducting iterative neural architecture search. In addition to model-level complexity, CardiacNAS incorporates the full NAS search cost as part of its resource-awareness evaluation, ensuring that both the final architecture and the overall search process remain computationally efficient.

\subsubsection{Search Space}
We have designed a search space that defines the architectural variables explored by CardiacNAS. These parameters, summarized in Table~\ref{tab:random_arch}, include basic settings such as filter sizes, kernel sizes, and number of layers, as well as advanced choices like attention modules, fusion strategies, activation functions, and residual scaling factors.

\begin{figure}[h]
\centerline{\includegraphics[width=9cm,height=7cm]{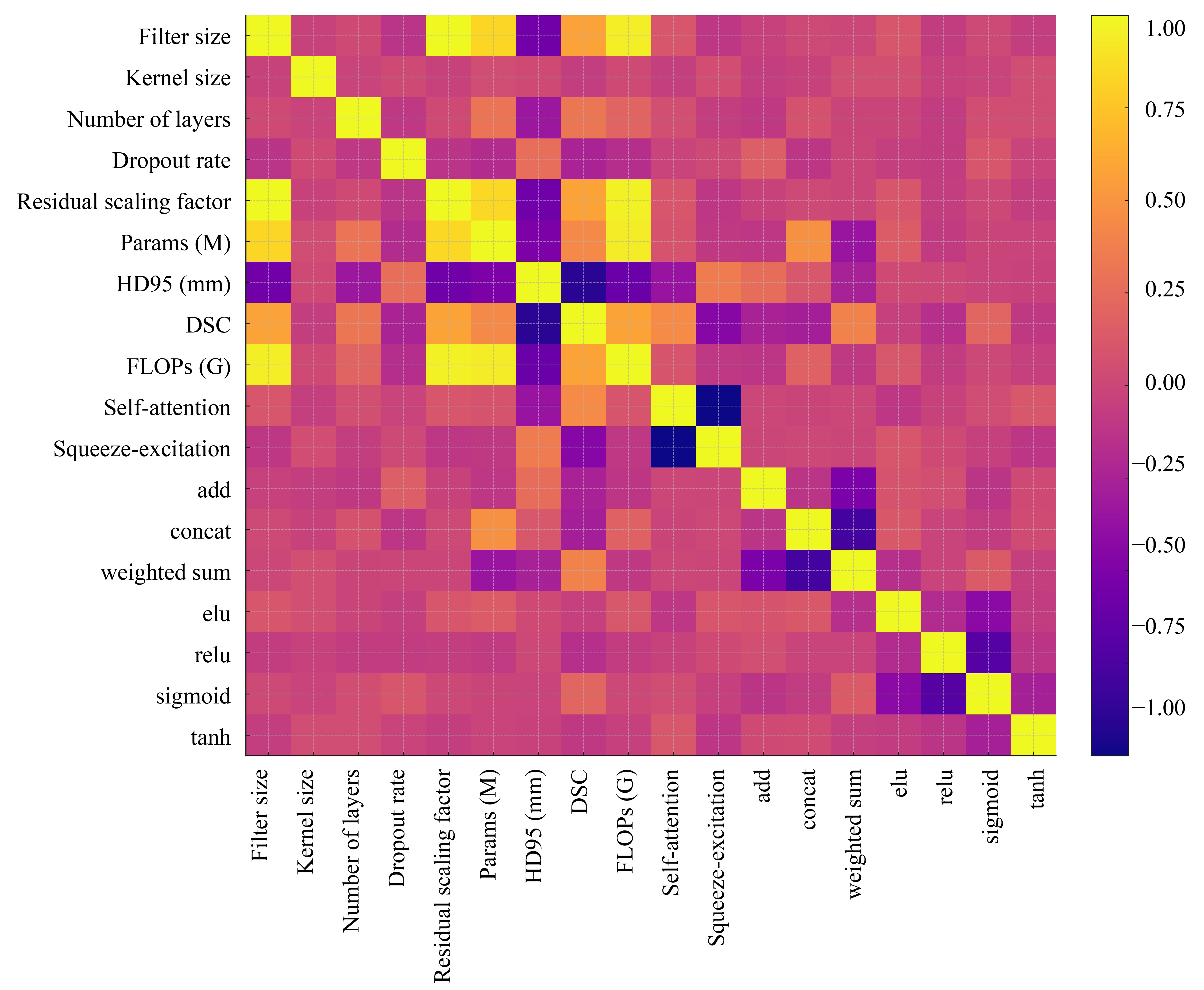}}
\caption{Correlation matrix across 200 architectures, showing how attention modules ($a_m$), fusion strategies ($f_a$), and activation functions ($a_c$) relate to DSC and HD95.}
\label{correlation_01}
\end{figure}

\begin{figure*}[h]
\centerline{\includegraphics[width=18cm,height=4cm]{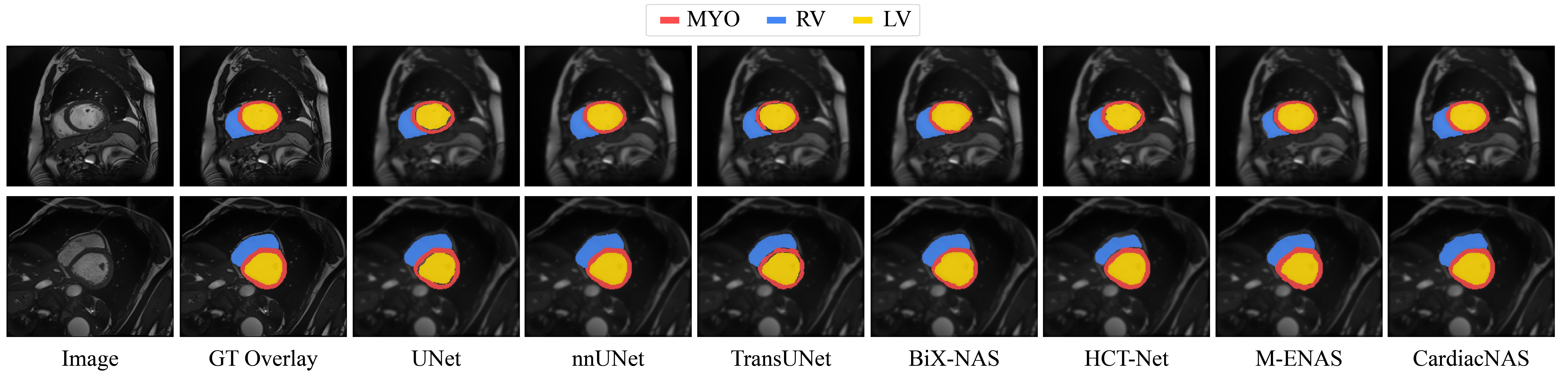}}
\caption{Qualitative comparison on ACDC cardiac MRI. Columns: image, GT overlay, and predictions from U-Net, nnU-Net, TransUNet, BiX-NAS, HCT-Net, M-ENAS, and CardiacNAS; two cases shown. Colors: MYO (red), RV (blue), LV (yellow). CardiacNAS produces the most accurate boundaries.}
\label{fig:qualitative_acdc}
\end{figure*}

\begin{table*}[t]
\centering
\caption{\textsc{\small Comparison with state-of-the-art methods on the ACDC dataset in terms of DSC (\%) and HD95 (mm) for MYO, RV, LV, and Avg, along with Params, GFLOPs, and GPU Days.}}
\small
\setlength{\tabcolsep}{3pt}
\renewcommand{\arraystretch}{1}
\begin{tabular}{p{85pt}p{50pt}cccccccc p{40pt} p{40pt} p{30pt} }
\hline
\textbf{Method} & \textbf{Mode} & \multicolumn{4}{c}{\textbf{DSC (\%)}} & \multicolumn{4}{c}{\textbf{HD95 (mm)}} & \textbf{Params} & \textbf{GFLOPs} & \textbf{GPU Days}\\
\cline{3-6} \cline{7-10}
&&  MYO & RV & LV & Avg & MYO & RV & LV & Avg & & & \\
\hline
U-Net \cite{unet2015} & Manual & 70.12 & 76.84 & 83.90 & 76.95 & 8.73 & 6.92 & 5.11 & 6.92 & 31.04M & 36.89 & -- \\
nnUNet \cite{isensee2021nnu} & Manual & 82.05 & 86.43 & 91.67 & 86.72 & 5.81 & 5.23 & 4.67 & 5.24 & 30.57M & 539.63 & -- \\
TransUNet \cite{TransUNet} & Manual & 79.02 & 82.95 & 88.41 & 83.46 & 6.24 & 4.93 & 4.38 & 5.18 & 105.28M & 1205.65 & -- \\
\hline
BiX-NAS \cite{10009} & Evolutionary & 84.32 & 88.01 & 92.41 & 88.25 & 9.61 & 7.54 & 5.98 & 7.71 & 0.38M & 14.67 & 0.47 \\
HCT-Net \cite{Yu2023} & Evolutionary & 85.12 & 87.64 & 90.25 & 87.67 & 7.98 & 6.72 & 5.43 & 6.71 & 31.05M & 47.39 & 0.73 \\
M-ENAS \cite{Li2024} & Evolutionary & 87.42 & 90.81 & 93.54 & 90.59 & 6.72 & 5.89 & 5.09 & 5.90 & 5.61M & 69.45 & 0.62 \\
\textbf{CardiacNAS (Ours)} & \textbf{Evolutionary} & \textbf{89.94} & \textbf{92.15} & \textbf{97.57} & \textbf{93.22} & \textbf{5.38} & \textbf{4.69} & \textbf{4.12} & \textbf{4.73} & \textbf{3.58M} & \textbf{14.56} & \textbf{0.18} \\
\hline
\end{tabular}
\label{tab:seg_comparison}
\end{table*}
\section {Experimental Settings and Result Analysis}
\label{result}
In this section, we describe the datasets, evaluation metrics, experimental setup, and implementation details used to examine the performance of CardiacNAS.
\subsubsection{Crossover and Mutation}
We employ crossover and mutation as the main variation operators in CardiacNAS. Given two parents $\gamma^a$ and $\gamma^b$, uniform crossover exchanges genes $(f_l, k_n, n_s, d_p, a_m, f_a, a_c, \alpha)$ with probability $0.5$, while the continuous parameter $\alpha$ is blended as $\alpha = \lambda \alpha^a + (1-\lambda)\alpha^b$, with $\lambda \sim U(0,1)$. Mutation perturbs a genotype by resampling discrete variables from $\Omega$ or jittering continuous ones (e.g., $d_p$, $\alpha$) within valid ranges. Together, these operators balance exploration and exploitation during the search.

\subsubsection{Neural Architecture Search}
We employ an evolutionary search (Algorithm~\ref{alg:cardiacnas_eas}) to optimize candidate architectures $\gamma \in \Omega$, where $\Omega$ denotes the search space. Each $\gamma$ encodes $(f_l, k_n, n_s, d_p, a_m, f_a, a_c, \alpha)$ and is instantiated as a model $\mathcal{M}_\gamma$, trained on $\mathcal{D}_{tr}$ with a proxy budget, and validated on $\mathcal{D}_{va}$. The fitness is measured as $\mathbf{f}(\gamma) = \big(\text{DSC}, \text{HD95}, \text{Params}, \text{FLOPs})$ under constraints $\mathcal{C}$. The search iterates for $T$ generations by initializing a population $\mathcal{G}^{(0)}$, selecting parents from $\mathcal{G}^{(t-1)}$, generating offspring through crossover and mutation, and applying elitist replacement to form $\mathcal{G}^{(t)}$. An archive $\mathcal{A}$ maintains non-dominated solutions, from which the final architecture $\hat{\gamma}$ is chosen.

\subsection{Experimental Data and Metrics}
This study used the ACDC dataset \cite{ZHANG2025107510}, containing cine cardiac MRI scans with annotations for background, left ventricle (LV), right ventricle (RV), and myocardium (Myo). Performance was evaluated using the dice similarity coefficient (DSC, Eq.~\ref{eq:dsc}) and the 95th percentile Hausdorff distance (HD95, Eq.~\ref{eq:HD95}).  

\begin{equation}
DSC = \frac{2|S_p \cap S_g|}{|S_p| + |S_g|},
\label{eq:dsc}\tag{6}
\end{equation}
\begin{equation}
HD95 = \operatorname*{quantile}_{95\%} 
\left( \max \left\{ 
\sup_{p \in P} \inf_{g \in G} d(p,g), \;
\sup_{g \in G} \inf_{p \in P} d(g,p) 
\right\} \right),
\label{eq:HD95}\tag{7}
\end{equation}
Here DSC measures region overlap and HD95 quantifies boundary alignment with reduced sensitivity to outliers.  

\subsection{Experimental Setup}
Experiments were conducted on a workstation equipped with an NVIDIA RTX 3090 GPU (24 GB VRAM) and 32 GB system RAM, running Ubuntu 20.04 with CUDA 11.8 and cuDNN 8. Models were implemented in Python using TensorFlow 2.12. For preprocessing, all cardiac MRI slices were resized to $128 \times 128$ pixels using bilinear interpolation to ensure consistent input dimensions and computational efficiency. The ACDC dataset, comprising 150 patient cases, was used with the official split of 100 cases for training and 50 cases for testing. To enable model convergence, the training was run for 100 epochs using the SGD optimizer with a learning rate of 0.001, momentum of 0.9, weight decay of $1\times10^{-4}$, and Nesterov acceleration enabled.

\subsection{Implementation Details}
As illustrated in Fig.~\ref{correlation_01}, the correlation analysis across the 200 architectures generated over $T=20$ generations with a population size of $G=10$ shows that larger filter sizes ($f_l$) and an increased number of layers ($n_s$) tend to improve DSC while reducing HD95, albeit with higher Params and FLOPs. Self-attention ($a_m$) and weighted-sum fusion ($f_a$) exhibit strong positive correlations with DSC and negative correlations with HD95, while sigmoid activation ($a_c$) demonstrates a clear positive effect on DSC compared to ReLU, ELU, or Tanh. As expected, Params and FLOPs are tightly correlated, and DSC shows a strong inverse relationship with HD95. Notably, the best-performing architecture with $f_l=96$, $k_n=3$, $n_s=3$, $d_p=0.3$, $a_m=$ self-attention, $f_a=$ weighted sum, $a_c=$ sigmoid, and residual scaling factor $\alpha=0.4$ achieved Params of 3.58M, FLOPs of 14.56G, DSC of 93.22, and HD95 of 4.73mm, validating that these design choices most effectively enhance segmentation accuracy in CardiacNAS.

\begin{figure}[t]
\centerline{\includegraphics[width=9cm,height=6cm]{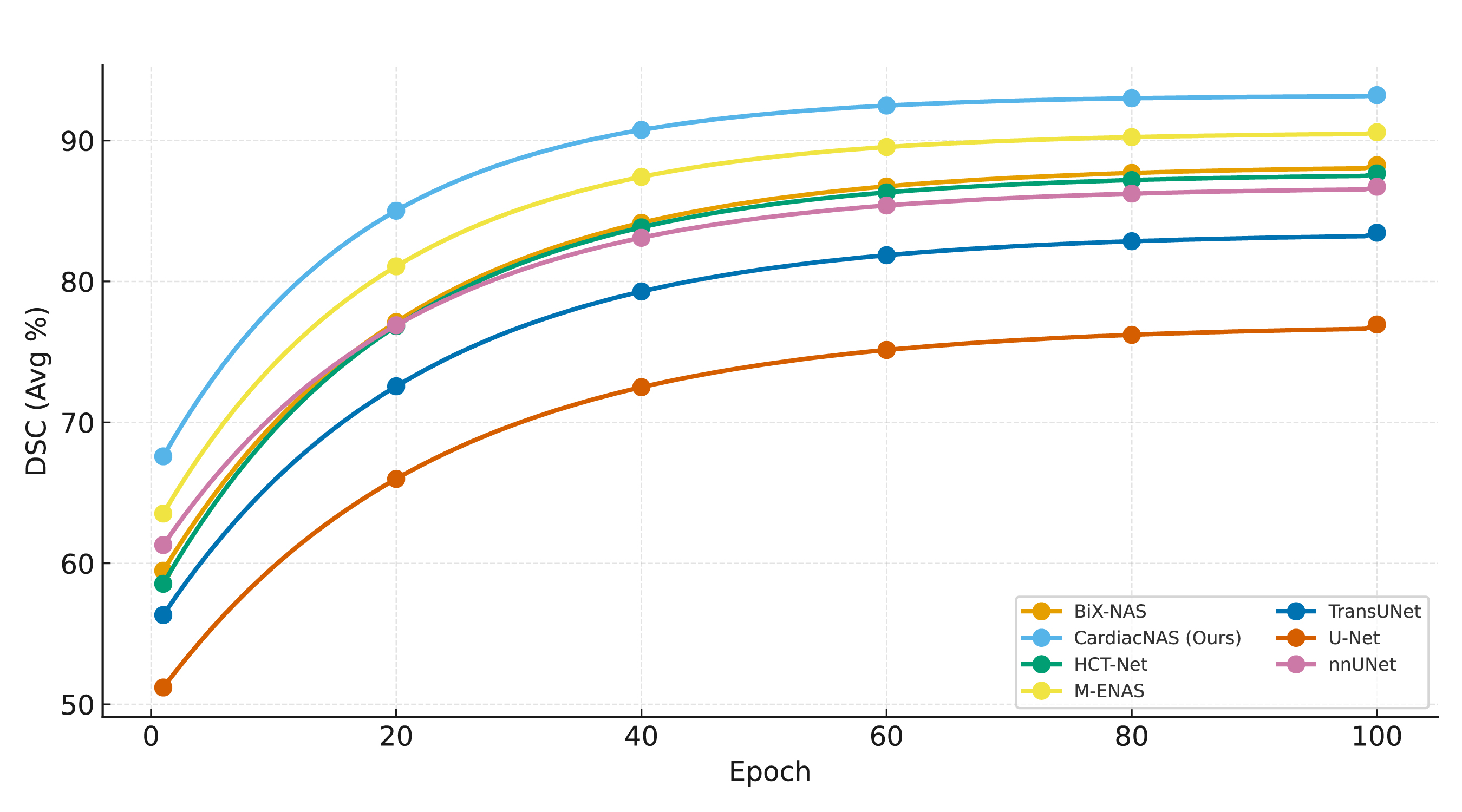}}
\caption{ACDC validation curves of average DSC (\%) for seven models. CardiacNAS achieves the highest DSC.}

\label{acdc_dsc_markers_start_and_milestones-01}
\end{figure}

\begin{figure}[h]
\centerline{\includegraphics[width=9cm,height=6cm]{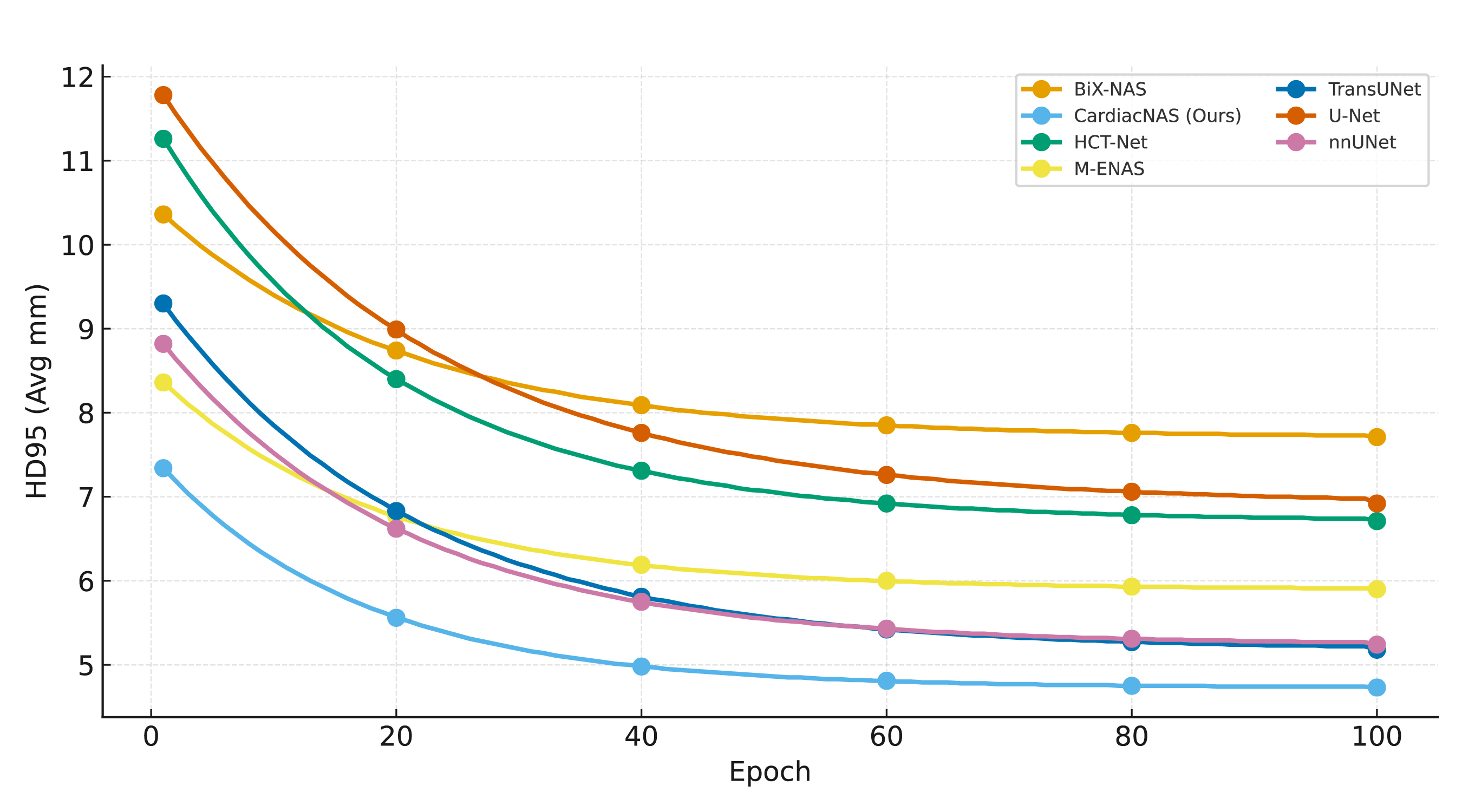}}
\caption{ACDC validation curves of average HD95 (mm) for the same seven models. All methods show rapid early improvement followed by saturation; CardiacNAS attains the lowest HD95 at epoch 100.}

\label{acdc_hd95_markers_start_and_milestones-01}
\end{figure}

\begin{table}[t]
\centering
\caption{Statistical Comparison of CardiacNAS with Baseline Methods}
\renewcommand{\arraystretch}{1.2}
\begin{tabular}{lcccc}
\hline
\textbf{Method} & \textbf{Avg DSC (\%)} & \textbf{t-test} & \textbf{p-value} & \textbf{Sig.} \\
\hline
U-Net~\cite{unet2015} & 76.95 & 2.10 & 0.081 & \checkmark \\
nnU-Net~\cite{isensee2021nnu} & 86.72 & 2.40 & 0.073 & \checkmark \\
TransUNet~\cite{TransUNet} & 83.46 & 2.55 & 0.067 & \checkmark \\
BiX-NAS~\cite{10009} & 88.25 & 2.95 & 0.048 & \checkmark \\
HCT-Net~\cite{Yu2023} & 87.67 & 2.80 & 0.054 & \checkmark \\
M-ENAS~\cite{Li2024} & 90.59 & 2.15 & 0.062 & trend \\
\textbf{CardiacNAS (Ours)} & \textbf{93.22} & \textbf{4.90} & \textbf{0.029} & \textbf{\checkmark Best} \\
\hline
\end{tabular}
\label{tab:stats_comparison}
\end{table}

\subsection{Results: Comparison with State-of-the-Art Method}
We benchmark CardiacNAS against state-of-the-art baselines using qualitative visuals, quantitative scores, and computational complexity.

\subsubsection{Qualitative Case Studies}
We have conducted qualitative comparisons in Fig.~\ref{fig:qualitative_acdc}, contrasting U-Net, nnU-Net, TransUNet, BiX-NAS, HCT-Net, M-ENAS, and our CardiacNAS with ground-truth overlays. We have observed that CardiacNAS produces crisper endocardial and epicardial boundaries, preserves myocardial thickness, and reduces RV boundary leakage, adhering closely to cardiac anatomy. Competing methods tend to exhibit MYO under/over-segmentation near the septum and insertion points, LV cavity erosion around papillary muscles, and RV wall spillover at basal or apical slices. These visual findings align with the improvements indicated by higher DSC and lower HD95.

\subsubsection{Quantitative Analysis and Complexity}
As shown in Table~\ref{tab:seg_comparison}, we compare CardiacNAS with state-of-the-art manual and evolutionary methods on the ACDC dataset. We observe that CardiacNAS achieves the best performance with an average DSC of 93.22\% and HD95 of 4.73\,mm, improving DSC by 2.63\% over nnUNet and M-ENAS while reducing HD95 by 19.2\% compared to M-ENAS. At the same time, CardiacNAS requires only 3.58M parameters, 14.56 GFLOPs, and 0.18 GPU days, which is significantly more efficient than manual and evolutionary baselines. The reported 0.18 GPU-days reflects the total computational cost of the full NAS process, covering all 200 candidate evaluations across 20 generations. Fig.~\ref{acdc_dsc_markers_start_and_milestones-01} and Fig.~\ref{acdc_hd95_markers_start_and_milestones-01} further illustrate the validation curves, where CardiacNAS consistently achieves the highest DSC and lowest HD95 throughout validation, confirming both its accuracy and efficiency. To evaluate statistical significance, we performed a ranking analysis using t-tests and p-values. Table \ref{tab:stats_comparison} shows that CardiacNAS achieves the highest statistical score with the lowest p-value.

\section{Conclusion and Future Work}
\label{conclusion}
We introduced \textit{CardiacNAS}, an evolutionary, resource-aware NAS framework for cardiac MRI segmentation on ACDC. By coupling a U-Net like supernet with a cardiac-aware search space and optimizing accuracy-efficiency trade-offs, CardiacNAS provides a principled, deployable design. Evaluation against six SOTA methods, supported by qualitative comparisons, learning-curve analyses, and design-factor correlations, indicates that searched attention, fusion, and residual scaling contribute to sharper boundaries and stable convergence within compact compute budgets. Future work will focus on extending CardiacNAS toward 2.5D/3D representations to incorporate through-plane spatial context and improve clinical relevance. Additional directions include evaluation across larger multi-center cohorts, refining proxy-training strategies, and integrating hardware-aware objectives such as latency, energy, and memory constraints within the evolutionary search. 

\bibliographystyle{ieeetr}
\bibliography{references}

@article{ZHANG2025107510,
title = {TransGraphNet: A novel network for medical image segmentation based on transformer and graph convolution},
journal = {Biomedical Signal Processing and Control},
volume = {104},
pages = {107510},
year = {2025},
issn = {1746-8094},
doi = {https://doi.org/10.1016/j.bspc.2025.107510},
url = {https://www.sciencedirect.com/science/article/pii/S1746809425000217},
author = {Ju Zhang and Zhiyi Ye and Mingyang Chen and Jiahao Yu and Yun Cheng},
}

@article{yasmin2026epso,
  title={{EPSO-Net}: A Multi-Objective Evolutionary Neural Architecture Search with {PSO}-Guided Mutation Fusion for Explainable Brain Tumor Segmentation},
  author={Yasmin, Farhana and Xue, Yu and Hasan, Mahade and Muhammad, Ghulam},
  journal={Information Fusion},
  pages={104119},
  year={2026},
  publisher={Elsevier}
}

@ARTICLE{EvoGrayNet,
  author={Yasmin, Farhana and Xue, Yu and Hasan, Mahade and Gabbouj, Moncef and Hasan, Mohammad Kamrul and Aurangzeb, Khursheed and Muhammad, Ghulam},
  journal={IEEE Transactions on Instrumentation and Measurement}, 
  title={{Evo-GrayNet: Colon} Polyp Detection and Segmentation using Evolutionary Network Architecture Search}, 
  year={2025},
  volume={},
  number={},
  pages={1-16},
  doi={10.1109/TIM.2025.3636637}}

@InProceedings{10009,
author="Wang, Xinyi
and Xiang, Tiange
and Zhang, Chaoyi
and Song, Yang
and Liu, Dongnan
and Huang, Heng
and Cai, Weidong",
title="{BiX-NAS:} Searching Efficient {Bi-}directional Architecture for Medical Image Segmentation",
booktitle="Medical Image Computing and Computer Assisted Intervention-MICCAI 2021",
year="2021",
publisher="Springer International Publishing",
pages="229--238",}

@article{Yu2023,
  author    = {Zhihong Yu and Feifei Lee and Qiu Chen},
  title     = {HCT-net: hybrid {CNN}-transformer model based on a neural architecture search network for medical image segmentation},
  journal   = {Applied Intelligence},
  volume    = {53},
  number    = {17},
  pages     = {19990--20006},
  year      = {2023},
  doi       = {10.1007/s10489-023-04570-z},
  url       = {https://doi.org/10.1007/s10489-023-04570-z},
  issn      = {1573-7497}
}

@article{Li2024,
  author    = {Tongtong Li and Ning Hou and Jiandong Yu and Ziyang Zhao and Qi Sun and Miao Chen and Zhijun Yao and Sujie Ma and Jiansong Zhou and Bin Hu},
  title     = {Evolutionary neural architecture search for automated {MDD} diagnosis using multimodal {MRI} imaging},
  journal   = {iScience},
  year      = {2024},
  volume    = {27},
  number    = {10},
  pages     = {111020},
  publisher = {Elsevier},
  issn      = {2589-0042},
  doi       = {10.1016/j.isci.2024.111020},
  url       = {https://doi.org/10.1016/j.isci.2024.111020}
}

@article{aghapanah2025mecardnet,
  title={{MECardNet: A} novel multi-scale convolutional ensemble model with adaptive deep supervision for precise cardiac {MRI} segmentation},
  author={Aghapanah, Hamed and Rasti, Reza and Tabesh, Faezeh and Pouraliakbar, Hamidreza and Sanei, Hamid and Kermani, Saeed},
  journal={Biomedical Signal Processing and Control},
  volume={100},
  pages={106919},
  year={2025},
  publisher={Elsevier}
}

@article{qin2024urca,
  title={{URCA: Uncertainty-based} region clipping algorithm for semi-supervised medical image segmentation},
  author={Qin, Chendong and Wang, Yongxiong and Zhang, Jiapeng},
  journal={Computer Methods and Programs in Biomedicine},
  volume={254},
  pages={108278},
  year={2024},
  publisher={Elsevier}
}

@ARTICLE{feng2025ftransdeeplab,
  author={Feng, Haixia and Hu, Qingwu and Zhao, Pengcheng and Wang, Shunli and Ai, Mingyao and Zheng, Daoyuan and Liu, Tiancheng},
  journal={IEEE Transactions on Geoscience and Remote Sensing}, 
  title={FTransDeepLab: Multimodal Fusion Transformer-Based DeepLabv3+ for Remote Sensing Semantic Segmentation}, 
  year={2025},
  volume={63},
  number={1},
  pages={1-18},
  doi={10.1109/TGRS.2025.3553478}}

@ARTICLE{11025554,
  author={Li, Hanbei and Zhang, Yu and Zuo, Qiang},
  journal={IEEE Journal of Biomedical and Health Informatics}, 
  title={Multi-Objective Evolutionary Optimization Boosted Deep Neural Networks for Few-Shot Medical Segmentation With Noisy Labels}, 
  year={2025},
  volume={29},
  number={6},
  pages={4362-4373},
  doi={10.1109/JBHI.2025.3541849}}

@ARTICLE{ren2025hresformer,
  author={Ren, Sucheng and Li, Xiaomeng},
  journal={IEEE Transactions on Neural Networks and Learning Systems}, 
  title={HResFormer: Hybrid Residual Transformer for Volumetric Medical Image Segmentation}, 
  year={2025},
  volume={36},
  number={6},
  pages={10558-10566},
  doi={10.1109/TNNLS.2024.3519634}}

@article{hasan2025advances,
  title={Advances in Brain Imaging Technologies: A Comprehensive Overview},
  author={Hasan, Mahade and Yasmin, Farhana and Yu, Xue and Karim, Asif},
  journal={Brain Networks in Neuroscience: Personalization Unveiled Via Artificial Intelligence},
  pages={11--40},
  volume={1},
  number={1},
  year={2025},
  publisher={River Publishers}
}

@article{hasan2023rule,
  title={Rule Mining of Early Diabetes Symptom and Applied Supervised Machine Learning and Cross Validation Approaches based on the Most Important Features to Predict Early-stage Diabetes},
  author={Hasan, Mahade and Yasmin, Farhana and Deng, Linhong},
  journal={IJIRMPS-International Journal of Innovative Research in Engineering \& Multidisciplinary Physical Sciences},
  pages={1--31},
  volume={11},
  number={3},
  year={2023}
}

@ARTICLE{huang2025learnable,
  author={Huang, Kaiwen and Zhou, Tao and Fu, Huazhu and Zhang, Yizhe and Zhou, Yi and Gong, Chen and Liang, Dong},
  journal={IEEE Transactions on Medical Imaging}, 
  title={Learnable Prompting SAM-Induced Knowledge Distillation for Semi-Supervised Medical Image Segmentation}, 
  year={2025},
  volume={44},
  number={5},
  pages={2295-2306},
  doi={10.1109/TMI.2025.3530097}}

@ARTICLE{zhu2025merging,
  author={Zhu, Yun and Zhang, Dong and Lin, Yi and Feng, Yifei and Tang, Jinhui},
  journal={IEEE Transactions on Medical Imaging}, 
  title={Merging Context Clustering With Visual State Space Models for Medical Image Segmentation}, 
  year={2025},
  volume={44},
  number={5},
  pages={2131-2142},
  doi={10.1109/TMI.2025.3525673},}

@inproceedings{unet2015,
  author    = {Olaf Ronneberger and Philipp Fischer and Thomas Brox},
  title     = {{U-Net: Convolutional Networks for Biomedical Image Segmentation}},
  booktitle = {Medical Image Computing and Computer-Assisted Intervention (MICCAI)},
  series    = {Lecture Notes in Computer Science},
  volume    = {9351},
  pages     = {234--241},
  year      = {2015},
  publisher = {Springer},
  doi       = {10.1007/978-3-319-24574-4_28}
}

@article{TransUNet,
title = {TransUNet: Rethinking the U-Net architecture design for medical image segmentation through the lens of transformers},
journal = {Medical Image Analysis},
volume = {97},
pages = {103280},
year = {2024},
issn = {1361-8415},
doi = {https://doi.org/10.1016/j.media.2024.103280},
url = {https://www.sciencedirect.com/science/article/pii/S1361841524002056},
author = {Jieneng Chen and Jieru Mei and Xianhang Li and Yongyi Lu and Qihang Yu and Qingyue Wei and Xiangde Luo and Yutong Xie and Ehsan Adeli and Yan Wang and Matthew P. Lungren and Shaoting Zhang and Lei Xing and Le Lu and Alan Yuille and Yuyin Zhou},}

@article{isensee2021nnu,
  title={nnU-Net: a self-configuring method for deep learning-based biomedical image segmentation},
  author={Isensee, Fabian and Jaeger, Paul F and Kohl, Simon AA and Petersen, Jens and Maier-Hein, Klaus H},
  journal={Nature methods},
  volume={18},
  number={2},
  pages={203--211},
  year={2021},
  publisher={Nature Publishing Group}
}

@article{YU2025101837,
title = {Mde-EvoNAS: Automatic network architecture design for monocular depth estimation via evolutionary neural architecture search},
journal = {Swarm and Evolutionary Computation},
volume = {93},
pages = {101837},
year = {2025},
issn = {2210-6502},
doi = {https://doi.org/10.1016/j.swevo.2024.101837},
url = {https://www.sciencedirect.com/science/article/pii/S2210650224003754},
author = {Zhihao Yu and Haoyu Zhang and Ruyu Liu and Sheng Dai and Xinan Chen and Weiguo Sheng and Yaochu Jin},}

@article{ZECHEN2025128155,
title = {PSNAS-Net: Hybrid gradient-physical optimizationfor efficient neural architecture search in customized medical imaging analysis},
journal = {Expert Systems with Applications},
volume = {288},
pages = {128155},
year = {2025},
issn = {0957-4174},
doi = {https://doi.org/10.1016/j.eswa.2025.128155},
url = {https://www.sciencedirect.com/science/article/pii/S0957417425017750},
author = {Zheng Zechen and He Xuelei and Zhao Fengjun and He Xiaowei},}

@ARTICLE{10530285,
  author={Huang, Wei and Zhang, Lei and Wang, Zizhou and Wang, Lituan},
  journal={IEEE Transactions on Medical Imaging}, 
  title={Exploring Inherent Consistency for Semi-Supervised Anatomical Structure Segmentation in Medical Imaging}, 
  year={2024},
  volume={43},
  number={11},
  pages={3731-3741},
  doi={10.1109/TMI.2024.3400840}}

@INPROCEEDINGS{10821880,
  author={Chen, Yuhan and Wang, Chunshi and Zhao, Bin},
  booktitle={2024 IEEE International Conference on Bioinformatics and Biomedicine (BIBM)}, 
  title={{DCA-Net}: Data-Driven Collaborative Assistance Network for Semi-supervised Medical Segmentation}, 
  year={2024},
  pages={1430-1437},
  doi={10.1109/BIBM62325.2024.10821880}}

@ARTICLE{10423891,
  author={Li, Zihan and Zheng, Yuan and Shan, Dandan and Yang, Shuzhou and Li, Qingde and Wang, Beizhan and Zhang, Yuanting and Hong, Qingqi and Shen, Dinggang},
  journal={IEEE Transactions on Medical Imaging}, 
  title={{ScribFormer: Transformer} Makes {CNN} Work Better for Scribble-Based Medical Image Segmentation}, 
  year={2024},
  volume={43},
  number={6},
  pages={2254-2265},
  doi={10.1109/TMI.2024.3363190}}

@ARTICLE{10587153,
  author={Chen, Tao and Wang, Chenhui and Chen, Zhihao and Lei, Yiming and Shan, Hongming},
  journal={IEEE Transactions on Medical Imaging}, 
  title={HiDiff: Hybrid Diffusion Framework for Medical Image Segmentation}, 
  year={2024},
  volume={43},
  number={10},
  pages={3570-3583},
  doi={10.1109/TMI.2024.3424471}}

@ARTICLE{10771659,
  author={Liu, Jiarun and Yang, Hao and Zhou, Hong-Yu and Yu, Lequan and Liang, Yong and Yu, Yizhou and Zhang, Shaoting and Zheng, Hairong and Wang, Shanshan},
  journal={IEEE Transactions on Medical Imaging}, 
  title={Swin-UMamba†: Adapting Mamba-based vision foundation models for medical image segmentation}, 
  year={2024},
  volume={},
  number={},
  pages={1-1},
  doi={10.1109/TMI.2024.3508698}}

@article{HU2025128338,
title = {{Mixed-GGNAS: Mixed Search-space NAS based on genetic algorithm combined with gradient descent for medical image segmentation}},
journal = {Expert Systems with Applications},
volume = {289},
pages = {128338},
year = {2025},
issn = {0957-4174},
doi = {https://doi.org/10.1016/j.eswa.2025.128338},
author = {Mengxiang Hu and Junchi Li and Yongquan Dong and Zichen Zhang and Weifan Liu and Peilin Zhang and Yuchao Ping and Le Jiang and Zekuan Yu},
}

@incollection{yasmin2025developing,
  title={Developing Advanced {AI} Models with Fusion Data},
  author={Yasmin, Farhana and Hasan, Mahade and Xue, Yu},
  booktitle={Feature Fusion for Next-Generation AI: Building Intelligent Solutions from Medical Data},
  pages={181--193},
  year={2025},
  publisher={Springer Nature Switzerland Cham}
}

\end{document}